\documentclass[11pt]{article}
\usepackage[margin=1in]{geometry}

\usepackage{amsmath}
\usepackage{amsfonts}
\usepackage{hyperref}
\usepackage{float}
\usepackage{caption}
\usepackage{graphicx}
\usepackage{subcaption}

\usepackage{xcolor}

\usepackage{soul}

\usepackage[american]{babel}

\usepackage{csquotes}
\usepackage[style=apa,sortcites=true,sorting=nyt,backend=biber]{biblatex}
\DeclareLanguageMapping{american}{american-apa} %
\usepackage[T1]{fontenc}

\title{Relational Schemata in BERT Are Inducible, Not Emergent:\\A Study of Performance vs. Competence in Language Models}

\author{Cole Gawin}

\begin{document}

\maketitle

\begin{abstract}

  While large language models like BERT demonstrate strong empirical performance on semantic tasks, whether this reflects true conceptual competence or surface-level statistical association remains unclear. I investigate whether BERT encodes abstract relational schemata by examining internal representations of concept pairs across taxonomic, mereological, and functional relations. I compare BERT's relational classification performance with representational structure in \texttt{[CLS]} token embeddings. Results reveal that pretrained BERT enables high classification accuracy, indicating latent relational signals. However, concept pairs organize by relation type in high-dimensional embedding space only after fine-tuning on supervised relation classification tasks. This indicates relational schemata are not emergent from pretraining alone but can be induced via task scaffolding. These findings demonstrate that behavioral performance does not necessarily imply structured conceptual understanding, though models can acquire inductive biases for grounded relational abstraction through appropriate training.
\end{abstract}

\section{Introduction}

Understanding how concepts relate to one another is essential to both human cognition and natural language understanding. From early developmental stages, humans categorize concepts and experiences not only by individual features, but also by how these concepts relate to one another \parencite{Gentner1983}. Importantly, this reasoning is not based solely on surface associations, but on abstract relational schemata: mental templates that encode the type of relationship connecting entities, independent of their specific identities \parencite{gentner2005relational}.

In recent years, large language models (LLMs) like BERT \parencite{devlin-etal-2019-bert} have achieved impressive performance on semantic and reasoning tasks, prompting interest in whether these models possess similar capacities for relational abstraction. However, there is growing evidence that strong performance on benchmarks does not necessarily imply conceptual competence \parencite{Bender2020, McCoy2019, Niven2019}.

Recent work from \textcite{Gawin2025} has shown that LLMs tend to default to \texttt{is-a} or other incorrect relations in open-ended relation classification tasks, even when other relation types are more appropriate, perhaps due to overrepresentation of these relation types in training data. This suggests a bias toward surface-level cues rather than structured conceptual abstraction.

These insights raise the question: do models like BERT naturally learn to group different kinds of relationships—like part-whole, category-membership, or function—into meaningful, abstract categories? In other words, is relational abstraction an emergent property of pretraining, or does it require explicit supervision through task scaffolding to induce?

To investigate this query, I examine BERT's behavior and internal representations on a relation classification task across different classes of semantic relations. I assess BERT's behavior on a relation classification task, and its innate ``knowledge'' or understanding through analysis of its internal representations. Critically, I compare these results before and after fine-tuning BERT on the classification task. If relational schemata are emergent properties of pretraining, it would be expected that concept pairs organize meaningfully even in the pretrained model. If not, and structure only appears after fine-tuning, this would suggest that relational abstraction must be induced through supervision.

The results reveal a striking dissociation: while BERT performs well on the classification task even without fine-tuning, its internal representations do not reflect meaningful relational organization until after being trained on the task. This supports the view that BERT's relational signals are decodable but not abstract—latent, but not structured—without explicit inductive guidance.

\section{Background \& Literature Review}

\subsection{Relational Reasoning}

Relational reasoning is the ability to understand how entities are connected, and is a central feature of human thought. It enables grounded commonsense reasoning, analogical inference, and causal understanding \parencite{davis2015commonsense, cattell1971abilities}.
Classic theories in cognitive science, such as \textit{structure-mapping theory} \parencite{Gentner1983}, argue that relational reasoning depends not just on recognizing that two entities are related, but on abstracting the \textit{type} of relation. Human cognition supports this by organizing knowledge into relational schemata—reusable, generalizable conceptual templates that encode structured relationships \parencite{gentner2005relational}.

There are many different classes of relations that serve different purposes. For example, \textbf{taxonomic relations} describe hierarchical or categorical relationships between concepts, such as ``a dog is an animal;'' \textbf{mereological relations} describe part-whole relationships, such as ``a wheel is part of a car;'' and \textbf{functional relations} describe the functional roles or causal relations between concepts, such as ``a stove is used for cooking.''

\subsection{Performance vs. Competence}

The distinction between possessing knowledge and the ability to employ it aligns with the long-standing distinction between linguistic \textit{competence} and \textit{performance} \parencite{Chomsky1965, Kaufer1979}. A system may exhibit correct behavior on a cognitive task (performance) without necessarily having structured internal representations that enable this behavior (competence). In language, this means a learner may incidentally produce grammatically correct sentences without representing underlying syntactic rules; in conceptual reasoning, it means a model may classify relations correctly without internalizing what those relations mean.

In the context of language models, performance can be measured through behavioral tasks (e.g., classification accuracy), while competence can be assessed through representational analyses of the model's internal embeddings.

\subsection{Internal Representations of Knowledge in Language Models}

Much of what we know about how language models represent knowledge comes from work on probing and representational analysis. For instance, \cite{tenney-etal-2019-bert} demonstrated that the internal architecture of language models like BERT model specific linguistic qualia. Moreover, \cite{liu-etal-2019-linguistic} found that BERT distributes and organizes knowledge across its layers: early layers capture morphological and syntactic features, middle layers encode phrase-level semantics, and deeper layers reflect task-specific abstractions. This suggests that BERT's internal representations are not monolithic, but rather organized hierarchically, with different layers capturing different levels of abstraction.

\subsection{Inducible vs. Emergent Structure}

A key question in understanding language models like BERT is whether their relational knowledge emerges naturally from unsupervised pretraining, or whether it must be explicitly induced through supervised learning. This distinction between emergent and inducible structure is central to ongoing debates in both machine learning and cognitive science.

\textbf{Emergent structure} refers to patterns that arise spontaneously during pretraining, without explicit supervision. \textcite{Mikolov2013} famously demonstrated that word embeddings trained on co-occurrence statistics could support vector arithmetic analogies (e.g., $\text{king} - \text{man} + \text{woman} \approx \text{queen}$), suggesting that some relational regularities can emerge from distributional patterns alone. \textcite{ethayarajh-2019-contextual} further shows that contextual embeddings generated by BERT may encode rich, context-sensitive information—but whether this structure reflects abstract, generalizable relational schemata remains an open question.

In comparison, \textbf{inducible structure} refers to schemata that become salient or organized only through task-specific fine-tuning or supervised training. \textcite{tenney-etal-2019-bert} show that BERT's internal representations become more linguistically aligned after fine-tuning, suggesting that relational and syntactic features are not fully formed in pretraining alone. Similarly, \textcite{merchant-etal-2020-happens} find that BERT’s ability to represent semantic relations improves substantially with fine-tuning, highlighting the importance of supervision in shaping conceptual organization. These findings resonate with cognitive science accounts that argue for the necessity of structured learning environments in supporting symbolic and relational reasoning \parencite{Lake2017}.

Together, this body of work motivates a deeper investigation into whether BERT encodes relational structure inherently, or only after task alignment—a distinction that has significant implications for how we interpret the model's internal representations.

\section{Theoretical Framework}

This study investigates whether BERT encodes abstract relational schemata—structured, conceptual representations of relationships—or merely captures surface-level associative cues that suffice for performing relation classification tasks. To explore this, I probe both BERT's behavior and internal representations using a multitude of approaches grounded in cognitive science, taking inspiration from computational linguistics and neuroscience.

A central conceptual distinction guiding this work is between \textit{performance} and \textit{competence}. While a model like BERT may excel at classifying semantic relations—for example, identifying ``wheel-car'' as a part-whole relation—this success does not necessarily imply that it understands the underlying relational structure. It might simply be leveraging distributional regularities from pretraining (e.g., the frequent co-occurrence of ``wheel'' and ``car'') without internalizing a generalizable concept of part-whole relations \parencite{Bender2020}. This study seeks to distinguish between these two levels of model ability.

\textbf{Performance} is defined behaviorally, as the ability of a classifier trained on BERT's embeddings to correctly predict the relation type between two concepts. To measure performance, I will assess the language model on a relation classification task; if the model performs well, it suggests at the very least that latent relational signals are present in the embeddings. However, behavioral analysis alone cannot determine whether those signals are abstractly represented within the model, only whether they are decodable.

\textbf{Competence} is defined representationally, as the presence of structured conceptual organization in BERT's internal representations—specifically, whether concept pairs that share the same relation type are embedded in similar ways. A model with relational competence should group together concept pairs that instantiate the same type of relation, regardless of the specific entities involved, in the high-dimensional embedding space. If BERT has relational competence, then concept pairs with the same relation should be geometrically close in the embedding space; in contrast, if BERT lacks relational competence, then concept pairs will be scattered, and any classification success will rely on surface-level features rather than structured schemas.

To test these predictions, I use representational similarity analysis (RSA) and dimensionality reduction methods to probe the internal structure of BERT's representations. RSA, originally developed in systems neuroscience, is a framework for characterizing and comparing the representational geometry of neural or model-derived activity patterns \parencite{Laakso2000, Kriegeskorte2008, Kriegeskorte2013}. It enables analysis of how conceptual information is organized within high-dimensional representational spaces by focusing on the relationships—i.e., the pattern of similarities and dissimilarities—between stimulus representations. Specifically, I compute representational dissimilarity matrices (RDMs) and assess how closely the geometry of BERT's embedding space aligns with a ground-truth similarity structure derived from known relational labels. This alignment is quantified by calculating the Spearman rank correlation coefficient \parencite{Spearman1904} between the model's RDM and the ground-truth target matrix. In doing so, RSA provides a principled way to evaluate whether the representational geometry of BERT's embeddings captures the abstract relational structure expected from the task.

This experimental framework hones in on a critical distinction between emergent and inducible relational structure. Emergent structure would manifest in BERT’s pretrained state, arising naturally from unsupervised exposure to language data. Inducible structure, by contrast, requires supervised fine-tuning to shape and organize internal representations according to relational categories. This distinction is central to interpreting whether relational abstraction is a natural consequence of language modeling or a product of explicit task alignment.

By probing both behavioral output and representational organization—before and after fine-tuning—this study aims to illuminate the degree to which relational knowledge in BERT is latent, structured, and generalizable, and whether its conceptual understanding is a byproduct of pretraining or a result of task-specific induction.

\section{Methodology}

This study evaluates whether BERT internalizes abstract relational schemata or merely encodes associative cues sufficient for relation classification. To address this, I designed a multi-part analysis that separately assesses performance and competence. Performance was measured behaviorally via classification accuracy on relation categories, while competence was assessed through representational similarity analysis (RSA) and visualization of embedding space geometry. By comparing BERT's behavior in both pretrained and fine-tuned conditions, I aimed to determine whether relational abstraction emerges naturally through pretraining, and if not, whether it could be induced through task-specific supervision.

\subsection{Classification Task}

The primary task involved in this study was a semantic relation classification problem: given a pair of concepts, the model must identify the high-level relational category that connects them. Concept pairs were grouped into three broad, cognitively motivated relational categories: \textit{taxonomic} relations, \textit{mereological} relations, and \textit{functional} relations.

Each input was formatted as a pair of concept tokens joined by a separator token, i.e. \texttt{``CONCEPT1 [SEP] CONCEPT2''}. The output was a categorical label indicating the relation type. The classification task was used to measure how well BERT's representations support task performance, as well as to provide a fine-tuning objective in later stages of the study.

\subsection{Dataset}

I constructed a custom dataset by extracting concept pairs from ConceptNet \parencite{speer2017conceptnet}, a large commonsense knowledge graph that includes labeled semantic relationships between concepts. I selected a balanced subset of concept pairs and their relation labels that map cleanly into the three relational categories described above. Specifically, 400 samples were chosen from nine relation labels (three from each of the aforementioned categories): \texttt{is-a}, \texttt{defined-as}, and \texttt{distinct-from} for taxonomic relations; \texttt{part-of}, \texttt{has-a}, and \texttt{made-of} for mereological relations; and \texttt{used-for}, \texttt{causes}, and \texttt{capable-of} for functional relations. This resulted in a dataset of 3600 concept pairs, with 1200 samples per relation type.

\subsection{Fine-Tuning Procedure}

To investigate the effects of fine-tuning on BERT's relational representations, I employed a supervised learning approach. The \texttt{bert-base-uncased} model from HuggingFace Transformers was used as the foundation model for the downstream fine-tuning task. The model was initialized with the pretrained weights from BERT, which had been trained on a large corpus of text data. This initialization provided a strong starting point for the model's understanding of language and semantic relationships.

Fine-tuning was conducted on the relation classification task, in which the model was trained to predict one of three high-level relation categories. The training process was conducted using the HuggingFace \texttt{Trainer} and \texttt{TrainingArguments} interfaces, and involved optimizing the model's parameters to minimize the cross-entropy loss between predicted and true labels. This procedure allowed for an assessment of whether BERT's internal representations could be organized to adhere to coherent relational schemata.

To evaluate the model's performance, I used a standard train-test split, where 70\% of the dataset was used for training and 30\% was reserved for testing. The test set was designed to include concept pairs that were disjoint from those in the training set, ensuring that the model's performance could be evaluated on unseen examples.

\subsection{Evaluation}

I evaluated the model before and after fine-tuning along two complementary axes: performance (behavioral success in relation classification) and competence (evidence of structured, relational representation in the internal embedding space). Performance was assessed through accuracy and confusion matrices on the relation classification task, while competence was assessed through structural analysis of the embedding space, using RSA and dimensionality reduction techniques.

\subsubsection{Task Performance Metrics}

To measure performance, I assessed both the pretrained and fine-tuned BERT models on the relation classification task. For the pretrained model, I trained a logistic regression classifier on the frozen BERT embeddings of the \texttt{[CLS]} token to predict relation categories. For the fine-tuned model, I used the in-model classifier that was trained during the fine-tuning process.

To measure performance, I assessed both the pretrained and fine-tuned BERT models on the relation classification task. The evaluation involved computing accuracy, precision, recall, and F1-score. Additionally, I calculated the Cohen's kappa score, a metric for inter-annotator agreement \parencite{Cohen1960, Artstein2008}, for both models to measure the agreement between predicted and true labels while accounting for chance agreement. This provided a more statistically grounded measure of performance. Altogether, these metrics provide insight into how well the model behaviorally distinguishes different types of relational meaning.

\paragraph{Control Condition}

To determine whether BERT's success on the classification task reflects meaningful relational signals, I conducted a control experiment in which relation labels were randomly permuted across concept pairs. This label-randomization destroys the semantic alignment between examples and their intended relation types while preserving the distribution of labels and input pairs.

A classifier trained on these randomized labels would be expected to perform at or near chance if BERT's embeddings do not encode spurious correlations that align with the permuted labels. If high accuracy were observed even in this condition, it would suggest the classifier is leveraging dataset artifacts rather than relational information. Hence, this control helps to verify that performance on the classification task with correct labels reflects genuine relational signals in the embeddings.

\subsubsection{Representational Similarity Analysis}

I perform representational similarity analysis (RSA) to evaluate the relational competence of BERT's embeddings—that is, whether relation types are encoded as structured, conceptual categories. In this context, RSA was used to examine whether the representational geometry of the embedding space reflects meaningful relational organization.

For a given set of concept pairs, I first computed a representational dissimilarity matrix (RDM) over the \texttt{[CLS]} embeddings. The RDM is a square matrix of size $N \times N$\footnote{Here, $N$ denotes the number of concept pairs used in the analysis. Each concept pair is associated with a relational category, and the $N \times N$ matrix captures the pairwise dissimilarity between their \texttt{[CLS]} embeddings.}, where each entry reflects the dissimilarity between the embeddings of two concept pairs. Specifically, I used one minus the cosine similarity as the dissimilarity metric, ensuring that higher values correspond to greater representational distance. A corresponding ground-truth RDM was constructed by assigning a dissimilarity of 0 to concept pairs sharing the same relation type and 1 to all others.

To quantify the degree of alignment between the model's representational geometry and the relational structure defined by the labels, I compute the Spearman rank correlation coefficient between the upper triangles of the model-derived and ground-truth RDMs. This correlation assesses the extent to which BERT's internal similarity structure reflects the expected categorical distinctions.

Taking inspiration from recent work by \textcite{gandikota2024evolution}, RSA was performed separately for the pretrained and fine-tuned BERT models, as well as for each of BERT's 13 hidden layers (including the embedding layer). This allows for tracking where in the network relational structure emerges and how it evolves through fine-tuning.

\subsubsection{Embedding Space Projections and Clustering Analysis}

To further investigate the structure of BERT's embeddings, I apply dimensionality reduction techniques to visualize the extent to which related embeddings cluster. Inspired by work from \textcite{Kriegeskorte2013} which introduced the concept of representational geometry to analyze both the content and format of neural codes, I explored how BERT organizes relational information in its high-dimensional embedding space.

Specifically, I projected the \texttt{[CLS]} token embeddings onto a two-dimensional plane using Uniform Manifold Approximation and Projection (UMAP; \cite{https://doi.org/10.48550/arxiv.1802.03426}), a dimensionality reduction technique that preserves local structure. These projections serve as visualizations of the model’s representational geometry, enabling intuitive inspection of clustering behavior according to relation type.

By comparing UMAP plots for both the pretrained and fine-tuned models, I assessed how fine-tuning modifies the geometry of the embedding space. These changes revealed how relational distinctions are encoded more explicitly post-training, reflecting a transformation in the internal representational geometry. These visualizations complement the RSA findings by providing a spatial interpretation of representational similarity, supporting the idea that representational geometry evolves to better support task-relevant abstraction and generalization.

\section{Results}

\subsection{Task Performance: Classification Accuracy}

I first evaluated the performance of BERT on the relation classification task. The results for inner-annotator agreement as measured by Cohen's kappa are summarized in \autoref{tab:cohen}, while detailed classification metrics are provided in \autoref{tab:classification}. Fine-tuned BERT model achieved a higher Cohen's kappa ($\kappa = 0.8528$) compared to the pretrained model ($\kappa = 0.6819$), indicating better agreement with the ground truth. Furthermore, pretrained BERT model achieved an accuracy of 79\% on the relation classification task, while the fine-tuned model achieved an accuracy of 90\%.

\medskip
\begin{table}[h]
  \centering
  \caption{Cohen's kappa ($\kappa$) scores for pretrained and fine-tuned BERT models on the relation classification task.}
  \label{tab:cohen}
  \begin{tabular}{|c|c|c|}
    \hline
    & \textbf{Pretrained} & \textbf{Fine-tuned} \\
    \hline
    \textbf{$\mathbf{\kappa}$} & 0.6819 & 0.8528 \\
    \hline
  \end{tabular}
\end{table}

These quanitative results, as well as the confusion matrices shown in \autoref{fig:sidebyside}, indicate that both pretrained and fine-tuned versions of BERT support accurate performance on relation classification. However, classifier success alone does not establish that relational categories are structurally represented within the model's embedding space. I turn to representational analyses to address this.

\subsection{Representational Competence: RSA and Embedding Structure}

To assess whether BERT organizes its internal representations according to abstract relational categories, I conducted representational similarity analysis (RSA) on \texttt{[CLS]} token embeddings from both pretrained and fine-tuned models. Spearman's rank correlation coefficient was computed for both models to quantify the degree of alignment between the model's classification predictions and ground-truth. This correlation quantifies how well the model's embeddings reflect the expected relational structure, with values closer to 1 indicating closer alignment. The results are summarized in \autoref{tab:spearman}.

\medskip
\begin{table}[H]
  \centering
  \caption{Spearman correlation ($\rho$) the model’s embedding similarity matrix and conceptual ground truth for pretrained and fine-tuned BERT models.}
  \label{tab:spearman}
  \begin{tabular}{|c|c|c|}
    \hline
    & \textbf{Pretrained} & \textbf{Fine-tuned} \\
    \hline
    \textbf{$\mathbf{\rho}$} & 0.0398 & 0.7706 \\
    \hline
  \end{tabular}
\end{table}

In the pretrained model, the Spearman coefficient was very low ($\rho \approx 0.04$) between the model's similarity matrix and the conceptual ground truth. After fine-tuning, the Spearman coefficient increased dramatically to $\rho \approx 0.77$, indicating a strong alignment between the model's internal representations and the expected relational structure.

\medskip
\begin{figure}[h]
  \centering
  \caption{Spearman correlation ($\rho$) between model layer representations and relational structure as measured by RSA, for each layer of the fine-tuned BERT model.}
  \includegraphics[width=\linewidth]{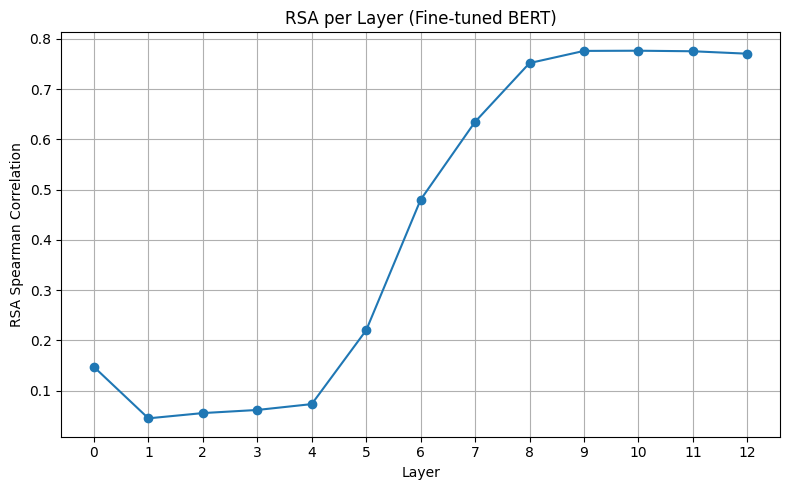}
  \label{fig:spearman-layerwise}
\end{figure}

RSA was also performed layer-wise for each of the fine-tuned model's 13 layers, including the embedding layer. As shown in \autoref{fig:spearman-layerwise}, the Spearman correlation coefficient increases dramatically in the deeper layers of the model. Correlation begins to dramatically increase in layer 5; by layer 8, the correlation exceeds $\rho = 0.7$, indicating that the model's internal representations are beginning to reflect the underlying relational structure. This trend continues through layers 9–12, where the Spearman correlation exceeds $\rho = 0.75$. This indicates that concept pairs sharing a relation category become consistently more similar to one another in embedding space.

\subsection{Visualization of Embedding Geometry}

\begin{figure}[h]
  \caption{UMAP projections of \texttt{[CLS]} token embeddings from pretrained (left) and fine-tuned (right) BERT models.}
  \centering
  \begin{subfigure}[b]{0.45\textwidth}
    \centering
    \includegraphics[width=\linewidth]{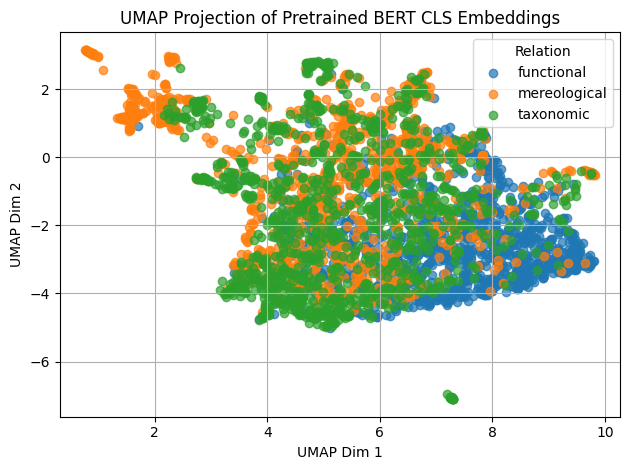}
  \end{subfigure}
  \hfill
  \begin{subfigure}[b]{0.45\textwidth}
    \centering
    \includegraphics[width=\linewidth]{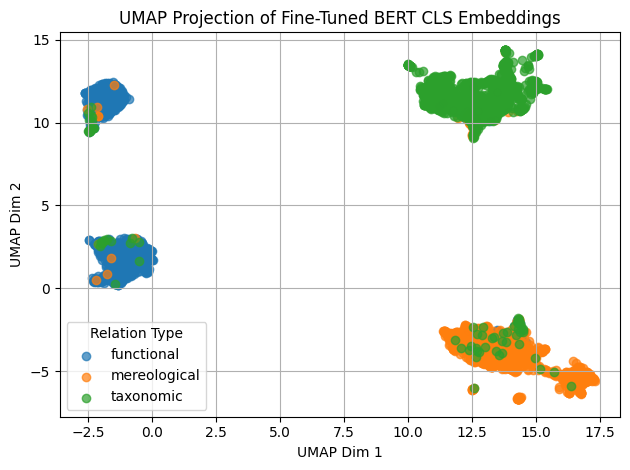}
  \end{subfigure}
  \label{fig:umap}
\end{figure}

To visualize the internal organization of relational knowledge, I projected \texttt{[CLS]} token embeddings into two dimensions using UMAP, both before and after fine-tuning. These visualizations are shown in \autoref{fig:umap}. The left plot shows the pretrained model, while the right plot shows the fine-tuned model. Each point represents a concept pair, colored by its relation category.

\medskip

In the pretrained model, no discernible clustering by relation category was observed. Concept pairs appeared interspersed, with substantial overlap across taxonomic, mereological, and functional classes.

In the fine-tuned model, a significant degree of relational clustering emerged. Concept pairs of the same relation type showed noticeably tighter grouping, and boundaries between categories became more apparent. There is still overlap between relation types, but the overall structure of the embedding space reflects the underlying relational categories. Intriguingly, the functional relation category appears to be split into two distinct clusters, while mereological and taxonomic relations are more tightly clustered together.

\subsection{Control Condition}

\begin{figure}[h]
  \centering
  \caption{Control condition results with randomized relation labels, showing confusion matrices (left) and UMAP projections on \texttt{[CLS]} token embeddings (right).}
  \begin{minipage}{0.45\textwidth}
    \centering
    \includegraphics[width=\linewidth]{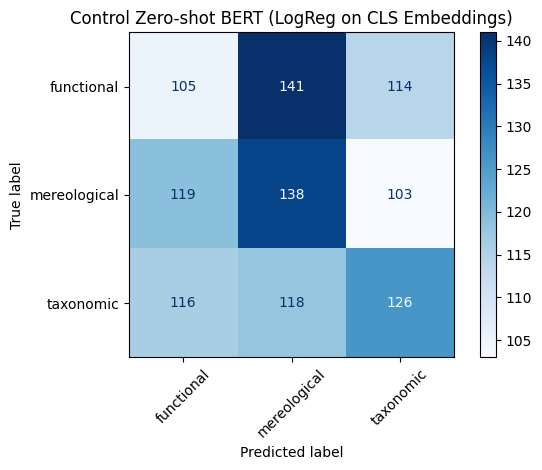}
  \end{minipage}
  \hspace{1cm} 
  \begin{minipage}{0.45\textwidth}
    \centering
    \includegraphics[width=\linewidth]{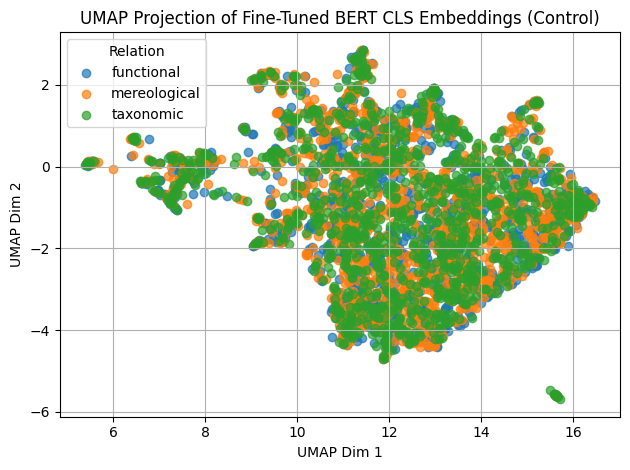}
  \end{minipage}
  \label{fig:control}
\end{figure}

To ensure that classification performance in the main condition reflects genuine relational signal rather than superficial statistical patterns, I implemented a control experiment in which relation labels were randomly permuted across concept pairs. Unlike the sharply diagonal confusion matrices observed in the non-randomized conditions shown in \autoref{fig:sidebyside}, the confusion matrix for the control condition in \autoref{fig:control} exhibits no clear predictive structure. Accuracy hovered near the expected chance level of one-third per class, with roughly equal misclassifications across all category combinations.

The right panel of \autoref{fig:control} displays a UMAP projection of \texttt{[CLS]} token embeddings for the pretrained BERT model trained on randomized labels. The projection reveals no observable clustering by relation type: concept pairs of all three categories are intermixed and diffusely distributed throughout the space. This lack of geometric structure contrasts sharply with the post-fine-tuning organization observed in the original task.

\medskip

Together, these control results support the validity of the study's main findings. They confirm that classification accuracy and RSA structure in the main condition are not artifacts of embedding geometry, label bias, or data leakage. Instead, they reflect real relational signals that are latent in pretrained BERT and become structurally organized only through fine-tuning.

\section{Discussion}

This study set out to test whether large language models like BERT internalize abstract relational schemata, or whether their ability to classify relations between concepts reflects surface-level associations. The findings reveal a striking dissociation between performance and competence in BERT’s relational behavior: while the model is capable of performing well on relation classification tasks, its internal representations do not initially reflect conceptual organization. Only after fine-tuning does BERT begin to exhibit representational structure consistent with relational competence.

Below, I discuss the implications of these findings for our understanding of BERT's relational knowledge and the nature of its internal representations.

\subsection{Performance Without Competence}

Before fine-tuning, a logistic regression classifier trained on frozen BERT embeddings achieved high accuracy in predicting relation categories. This indicates that latent relational signals are present and decodable in BERT's representations. However, deeper analysis using representational similarity analysis and dimensionality reduction revealed that these signals are not structurally organized. Concept pairs that shared the same relation type were no more similar to each other than to pairs from different categories, suggesting the absence of any schema-level structure.

This pattern highlights a \textit{performance-competence gap}: BERT can behave as if it understands the difference between different relational classes, but this performance arises from distributed associative cues, not from conceptual representations. The model succeeds because it has learned contextual signals associated with different relation types, not because it has abstracted those relations into structured categories.

\subsection{Limits of Co-occurrence-Based Reasoning}

This performance-competence gap observed in BERT echoes long-standing critiques in cognitive science (e.g., \cite{Fodor1988}), which argue that connectionist models (i.e., deep artificial neural networks) struggle to represent systematic, compositional, and relational structure. The findings here provide concrete empirical support for this critique in the context of modern LLMs: conceptual knowledge does not emerge from co-occurrence alone.

This raises important questions about the limits of co-occurrence-based reasoning in language models. While BERT can leverage statistical patterns to achieve high accuracy on relation classification tasks, this does not imply that it possesses a deep understanding of the underlying relational structure. Indeed, the model's performance on grounded reasoning tasks is presumably based on its ability to recognize and exploit surface-level cues, not on an abstract representation of the conceptual relations themselves \parencite{Bender2020, McCoy2019, Niven2019}.

\subsection{Latent Knowledge Is Not Structured Knowledge}

These findings suggest that BERT's relational knowledge is latent—meaning it can be decoded with appropriate supervision—but unstructured, in that it is not internally organized according to relation type. The model can leverage statistical cues to classify relations, but it does not represent those relations in a coherent, abstract form. The pretrained BERT model clearly ``knows'' that certain concept pairs are related in systematic ways, as evidenced by its high classification performance. However, this knowledge is latent and unstructured. It does not form a generalizable representational schema in the embedding space; instead, the model learns to associate specific cues in the input with output labels—association without abstraction. This is evident from the low RSA scores and the lack of clustering in the embedding space before fine-tuning.

This distinction has practical implications. In tasks where structural reasoning is essential—such as analogical inference, causal reasoning, or compositional generalization—models trained only on co-occurrence will likely fail. High accuracy on a fixed task does not guarantee that the model has internalized the conceptual structures needed for flexible, human-like reasoning.

However, the results also offer hope. Through fine-tuning, BERT was able to induce relational schemata, suggesting that co-occurrence provides a substrate, but not a scaffold. To move beyond associative reasoning, models must be shaped by explicit relational tasks, inductive biases, or architectures designed for structure.

\subsection{Inducing Conceptual Structure Through Fine-Tuning}

After fine-tuning BERT on the relation classification task, a striking change occurred: RSA scores rose sharply, particularly in the middle-to-deep layers (7–12).  Concept pairs belonging to the same relational category became geometrically clustered in embedding space. This aligns with previous work from \textcite{liu-etal-2019-linguistic}, which found that the later layers of BERT are more sensitive to task-specific scaffolding (i.e., relation classification). These findings are strong evidence that fine-tuning can induce abstract conceptual organization in BERT's internal representations—moving from unstructured, associative signals to structured, conceptually meaningful representations.

This transformation highlights that relational abstraction is not an emergent property of pretraining. Despite BERT's exposure to vast amounts of text, it does not spontaneously organize its knowledge by relation type. However, when exposed to a supervised task that scaffolds relational categories, it reorganizes its internal space to reflect that structure. This finding mirrors cognitive theories of human learning, where relational categories often require guided attention, explicit examples, and task constraints to emerge (e.g., \cite{Gentner1983}).

This parallel suggests that, like humans, BERT requires scaffolding to develop structured relational knowledge. In cognitive development, children often need repeated exposure, attention direction, and explicit task framing to form abstract relational categories. Similarly, BERT's relational competence does not emerge from passive exposure to linguistic co-occurrence during pretraining, but only when task-specific supervision forces the model to reorganize its internal representations along conceptual lines.

\section{Conclusion}

This study demonstrates a clear separation between performance and competence in BERT's handling of relational reasoning. While BERT’s pretrained embeddings support above-chance classification accuracy for semantic relations, this behavioral success does not reflect structured conceptual understanding. Through representational similarity analysis and embedding space visualization, it becomes evident that BERT’s internal geometry lacks relational schemata until fine-tuned on a supervised classification task. Only with task-specific scaffolding does the model's representational space begin to meaningfully organize according to abstract relation types. These findings suggest that relational abstraction is not an emergent property of pretraining but must be actively induced through guided learning. This distinction has important implications for interpreting language model capabilities and highlights the need for explicit structural biases or inductive signals when developing models intended for conceptual and relational reasoning tasks.

Future research should explore whether similar patterns of inducibility extend to other language models, including more recent architectures like LLaMa \parencite{https://doi.org/10.48550/arxiv.2302.13971} and Gemma \parencite{gemmateam2024gemmaopenmodelsbased}, and whether model scale influences the emergence of relational structure. Additionally, extending the range of tested relations to include \textit{temporal} or \textit{causal} relations could provide deeper insight into the scope of relational abstraction achievable in these models. Finally, bridging insights from cognitive development—such as curriculum learning or relational priming—could inspire new training paradigms that better align language model learning with human-like abstraction capabilities.

Ultimately, this work underscores the importance of task scaffolding in shaping the internal representations of language models, and suggests that relational abstraction is a learnable skill that can be induced through thoughtful training design.

\printbibliography

\newpage
\clearpage

\appendix
\section{Additional Figures}

\medskip
\begin{table}[ht]
  \centering
  \caption{Classification reports for pretrained (left) and fine-tuned (right) BERT models on the relation classification task.}
  \label{tab:classification}
  \begin{tabular}{|l||ccc||ccc|}
    \hline
    \textbf{Class} & \textbf{Precision} & \textbf{Recall} & \textbf{F1-score} & \textbf{Precision} & \textbf{Recall} & \textbf{F1-score} \\
    \hline
    Functional     & 0.84 & 0.85 & 0.84 & 0.92 & 0.95 & 0.94 \\
    Mereological   & 0.78 & 0.76 & 0.77 & 0.90 & 0.88 & 0.89 \\
    Taxonomic      & 0.75 & 0.75 & 0.75 & 0.89 & 0.88 & 0.88 \\
    \hline
    Accuracy       & \multicolumn{2}{c}{} & 0.79 & \multicolumn{2}{c}{} & 0.90 \\
    Macro avg      & 0.79 & 0.79 & 0.79 & 0.90 & 0.90 & 0.90 \\
    Weighted avg   & 0.79 & 0.79 & 0.79 & 0.90 & 0.90 & 0.90 \\
    \hline
  \end{tabular}
\end{table}

\medskip
\begin{figure}[ht]
  \centering
  \caption{Confusion matrices for relation classification using pretrained BERT with logistic regression on CLS embeddings (left) and fine-tuned BERT (right).}
  \begin{subfigure}[b]{0.45\textwidth}
    \centering
    \includegraphics[width=\linewidth]{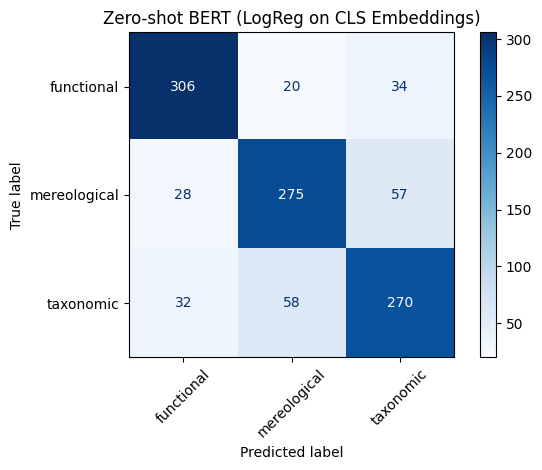}
    \label{fig:imageA}
  \end{subfigure}
  \hfill
  \begin{subfigure}[b]{0.45\textwidth}
    \centering
    \includegraphics[width=\linewidth]{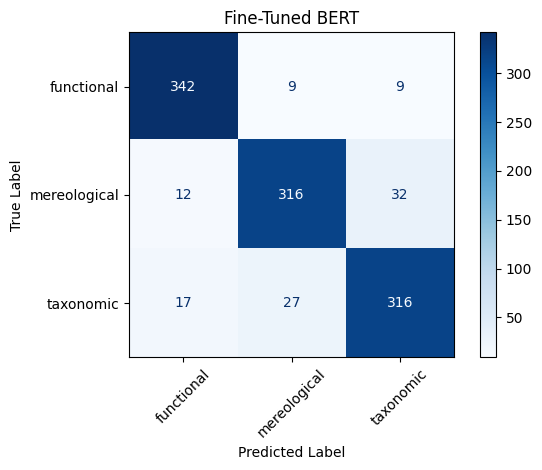}
    \label{fig:imageB}
  \end{subfigure}
  \label{fig:sidebyside}
\end{figure}

\end{document}